\documentclass[sigconf,nonacm]{acmart}

\usepackage{amsfonts}
\usepackage{amsmath}
\usepackage[inline]{enumitem}
\usepackage{graphicx}
\usepackage{hyperref}
\usepackage{listings}
\usepackage{multirow}
\usepackage{subcaption}
\usepackage{url}
\usepackage{wrapfig}
\usepackage{xcolor}

\copyrightyear{2022}
\acmYear{2022}
\setcopyright{acmcopyright}
\acmConference[SIGSPATIAL '22]{The 30th International Conference on Advances in Geographic Information Systems}{November 1--4, 2022}{Seattle, WA, USA}
\acmBooktitle{The 30th International Conference on Advances in Geographic Information Systems (SIGSPATIAL '22), November 1--4, 2022, Seattle, WA, USA}
\acmPrice{15.00}
\acmDOI{10.1145/3557915.3560953}
\acmISBN{978-1-4503-9529-8/22/11}

\begin{document}

\title{TorchGeo: Deep Learning With Geospatial Data}

\author{Adam J. Stewart}
\affiliation{%
    \institution{University of Illinois at Urbana-Champaign}
    \department{Department of Computer Science}
    \streetaddress{201 North Goodwin Avenue MC 258}
    \city{Urbana}
    \state{IL}
    \postcode{61801}
    \country{USA}
}
\email{adamjs5@illinois.edu}

\author{Caleb Robinson}
\affiliation{%
    \institution{Microsoft}
    \department{AI for Good Research Lab}
    \streetaddress{One Microsoft Way}
    \city{Redmond}
    \state{WA}
    \postcode{98052}
    \country{USA}
}
\email{caleb.robinson@microsoft.com}

\author{Isaac A. Corley}
\affiliation{%
    \institution{University of Texas at San Antonio}
    \department{Department of Electrical Engineering}
    \streetaddress{One UTSA Circle}
    \city{San Antonio}
    \state{TX}
    \postcode{78249}
    \country{USA}
}
\email{isaac.corley@my.utsa.edu}

\author{Anthony Ortiz}
\affiliation{%
    \institution{Microsoft}
    \department{AI for Good Research Lab}
    \streetaddress{One Microsoft Way}
    \city{Redmond}
    \state{WA}
    \postcode{98052}
    \country{USA}
}
\email{anthony.ortiz@microsoft.com}

\author{Juan M. Lavista Ferres}
\affiliation{%
    \institution{Microsoft}
    \department{AI for Good Research Lab}
    \streetaddress{One Microsoft Way}
    \city{Redmond}
    \state{WA}
    \postcode{98052}
    \country{USA}
}
\email{jlavista@microsoft.com}

\author{Arindam Banerjee}
\affiliation{%
    \institution{University of Illinois at Urbana-Champaign}
    \department{Department of Computer Science}
    \streetaddress{201 North Goodwin Avenue MC 258}
    \city{Urbana}
    \state{IL}
    \postcode{61801}
    \country{USA}
}
\email{arindamb@illinois.edu}

\renewcommand{\shortauthors}{Stewart et al.}

\begin{abstract}
Remotely sensed geospatial data are critical for applications including precision agriculture, urban planning, disaster monitoring and response, and climate change research, among others. Deep learning methods are particularly promising for modeling many remote sensing tasks given the success of deep neural networks in similar computer vision tasks and the sheer volume of remotely sensed imagery available. However, the variance in data collection methods and handling of geospatial metadata make the application of deep learning methodology to remotely sensed data nontrivial. For example, satellite imagery often includes additional spectral bands beyond red, green, and blue and must be joined to other geospatial data sources that can have differing coordinate systems, bounds, and resolutions. To help realize the potential of deep learning for remote sensing applications, we introduce TorchGeo, a Python library for integrating geospatial data into the PyTorch deep learning ecosystem. TorchGeo provides data loaders for a variety of benchmark datasets, composable datasets for generic geospatial data sources, samplers for geospatial data, and transforms that work with multispectral imagery. TorchGeo is also the first library to provide pre-trained models for multispectral satellite imagery (e.g., models that use all bands from the Sentinel-2 satellites), allowing for advances in transfer learning on downstream remote sensing tasks with limited labeled data. We use TorchGeo to create reproducible benchmark results on existing datasets and benchmark our proposed method for preprocessing geospatial imagery on the fly. TorchGeo is open source and available on GitHub: \url{https://github.com/microsoft/torchgeo}.
\end{abstract}



\maketitle

\section{Introduction} \label{sec:introduction}

\begin{figure*}[t]
    \centering
    \includegraphics[width=0.8\linewidth]{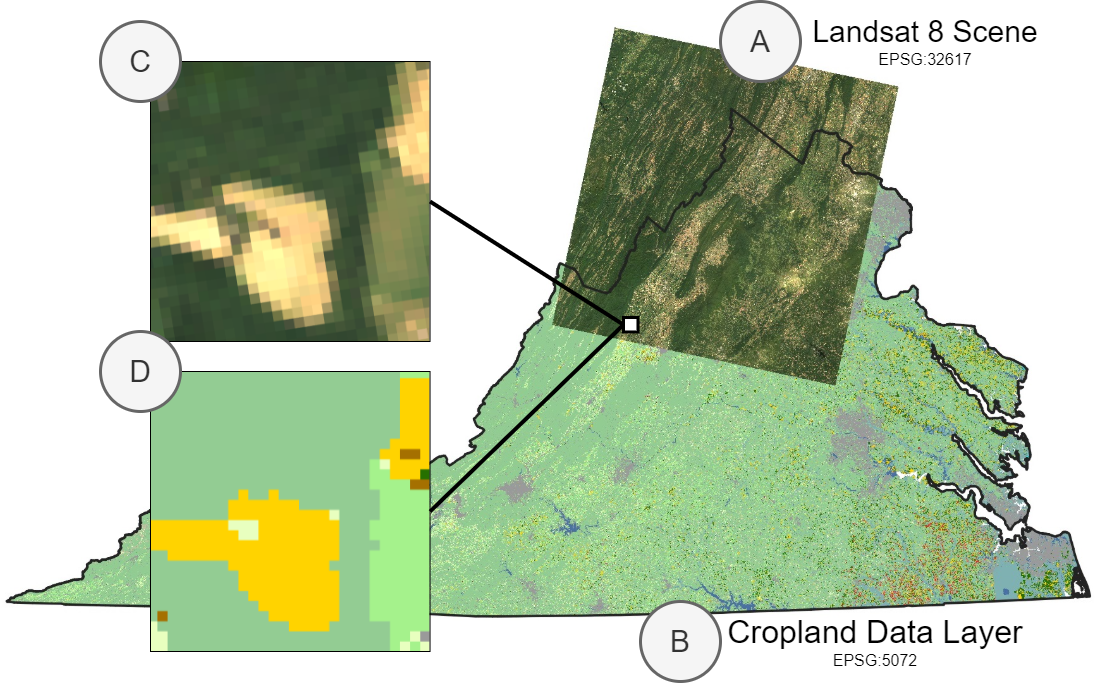}
    \caption{An illustration of the challenges in sampling from heterogeneous geospatial data layers. (\textbf{A} and \textbf{B}) show example geospatial data layers that a user may want to sample pixel-aligned data from. As these layers have differing coordinate reference systems, patches of imagery (\textbf{C} and \textbf{D}) sampled from these layers that cover the same area will not be pixel-aligned. TorchGeo transparently performs the appropriate alignment steps (reprojecting and resampling) during data loading such that users can train deep learning models without having to manually align the data layers.}
    \label{fig:main}
\end{figure*}

With the explosion in availability of satellite and aerial imagery over the past decades, there has been increasing interest in the use of imagery in remote sensing (RS) applications. These applications range from precision agriculture~\citep{MULLA2013358} and forestry~\citep{lu2016survey}, to natural and man-made disaster monitoring~\citep{van2013remote}, to weather and climate change~\citep{rolnick2019tackling}. At the same time, advancements in machine learning (ML), larger curated benchmark datasets, and increased compute power, have led to great successes in domains like computer vision, natural language processing, and audio processing. However, the wide-spread success and popularity of machine learning---particularly of deep learning methods---in these domains has not fully transferred to the RS domain, despite the existence of petabytes of freely available satellite imagery and a variety of benchmark datasets for different RS tasks. This is not to say that there are not successful applications of ML in RS, but that the full potential of the intersection of these fields has not been reached. Indeed, a recent book by~\citet{camps2021deep} thoroughly details work at the intersection of deep learning, geospatial data, and the Earth sciences. Increasing amounts of research on self-supervised and unsupervised learning methods specific to remotely sensed geospatial imagery~\citep{manas2021seasonal,ayush2021geography,jean2019tile2vec} bring the promise of developing generic models that can be tuned to various downstream RS tasks. Recent large-scale efforts, such as the creation of a global 10~m resolution land cover map~\citep{karra2021global} or the creation of global 30~m forest maps~\citep{rogan2008mapping}, pair the huge amount of available remotely sensed imagery with modern GPU accelerated models. To reach the full joint potential of these fields, we believe that we need tools for facilitating research and managing the complexities of both geospatial data and modern machine learning pipelines. We describe the challenges of this below, and detail our proposed solution, TorchGeo.

One major challenge in many RS tasks is the large amount of diversity in content of geospatial imagery datasets compared to datasets collected for traditional vision applications. For example, most conventional cameras capture 3-channel RGB imagery, however most satellite sensors capture different sets of spectral bands. The Landsat 8 satellite~\citep{roy2014landsat} collects 11 bands, the Sentinel-2 satellites~\citep{drusch2012sentinel} collect 12 bands, and the Hyperion satellite~\citep{pearlman2003hyperion} collects 242 (hyperspectral) bands, each measuring different regions of the electromagnetic spectrum. The exact wavelengths of the electromagnetic spectrum captured by each band can range from 400~nm to 15~$\mu$m. In addition, different sensors capture imagery at different spatial resolutions: satellite imagery resolution can range from 4~km/px (GOES~\citep{goes1994}) to 30~cm/px (Maxar WorldView satellites~\citep{maxar2021}), while imagery captured from drones can have a resolution as high as 7~mm/px~\citep{wingtra2021}. Depending on the type of orbit a satellite is in, imagery can be continuous (for geostationary orbits) or daily to biweekly (for polar, sun-synchronous orbits). Machine learning models or algorithms developed for one of these platforms will not generalize across inputs collected by the others, and, as a consequence of this, it is not possible to publish a single set of pre-trained model weights that span imaging platforms. In contrast, ImageNet~\citep{deng2009imagenet} pretrained models have been proven to be useful in a large number of transfer learning tasks~\citep{yosinski2014transferable}. Researchers and practitioners can often start with ImageNet pre-trained models in a transfer learning setup when presented with vision problems to reduce the overall amount of training needed to solve the problem. Further, it is not clear whether the inductive biases built into common modeling approaches for vision problems are immediately applicable to remotely sensed imagery. Large neural architecture search efforts~\citep{liu2018progressive} produce models that are optimized for and outperform hand-designed architectures on vision tasks, but it is an open question whether these transfer to remotely sensed imagery.

\begin{figure*}[t]
    \centering
    \includegraphics[width=1\linewidth]{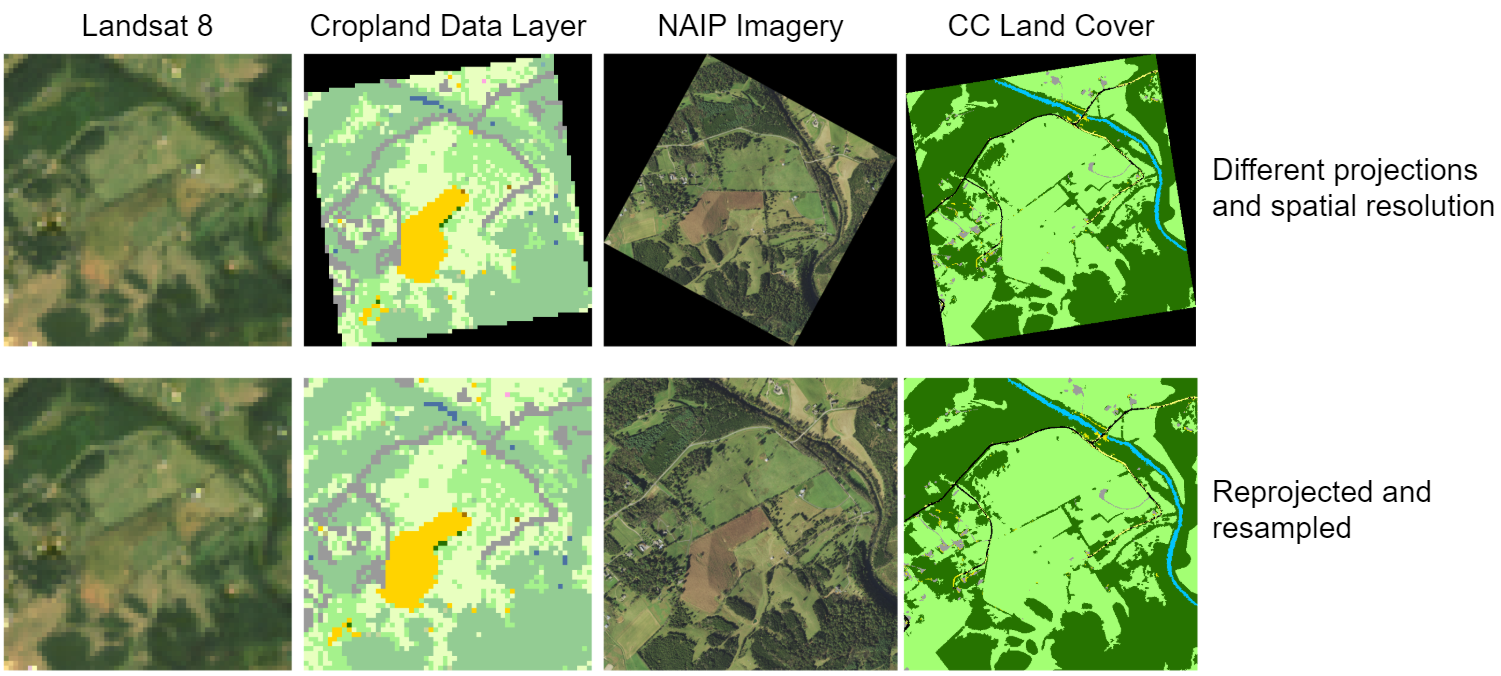}
    \caption{Different layers of geospatial data often have differing coordinate reference systems and spatial resolutions. (\textbf{Top row}) The same physical area cropped from four raster layers with different coordinate reference systems and spatial resolutions---this data is not pixel-aligned and cannot yet be used in modelling pipelines. (\textbf{Bottom row}) The same data after reprojecting into the same coordinate system and resampling to the highest spatial resolution---this data is pixel aligned and can serve as inputs or masks to deep neural networks.}
    \label{fig:warping-example}
\end{figure*}

Most machine learning libraries have not been designed to work with geospatial data. For example, the Python Imaging Library (PIL)~\citep{clark2015pillow}, used by many libraries to load images and perform data augmentation, does not support multispectral imagery. Similarly, deep learning models implemented by the torchvision library only support 3 channel (RGB) inputs, and must be adapted or re-implemented to support multispectral data. Datasets of geospatial data can be made up of a heterogenous mix of files with differing file formats, spatial resolutions, projections, and coordinate reference systems (CRS). Libraries such as GDAL~\citep{gdal2022} can interface with most types of geospatial data, however further abstractions for using such data in arbitrary deep learning pipelines are limited. Indeed, the gap between loading geospatial data from disk, and using it in a modeling pipeline, is large for all of the reasons mentioned above. As illustrated in Figure~\ref{fig:main}, users will often need pixel-aligned crops from multiple layers of data: imagery from different points in time over the same space, imagery and corresponding label masks, high-resolution and low resolution imagery from the same space, etc. In contrast, there are a wide variety of software libraries at the intersection of machine learning and other domains. In the PyTorch ecosystem, torchvision~\citep{torchvision}, torchtext~\citep{torchtext}, and torchaudio~\citep{torchaudio} provide the tools necessary for abstracting the peculiarities of domain-specific data away from the details of deep learning training pipelines.

To address these challenges, we propose TorchGeo, a Python package that allows users to transparently use heterogenous geospatial data in PyTorch-based deep learning pipelines. Specifically, TorchGeo provides:
\begin{enumerate}
    \item data loaders for geospatial datasets common in the literature,
    \item data loaders for combining uncurated geospatial raster and vector data layers with the ability to sample pixel-aligned patches on the fly,
    \item augmentations appropriate for multispectral imagery,
    \item data samplers appropriate for geospatial data, and
    \item pre-trained models for many common remotely sensed imagery sources. 
\end{enumerate}
In this paper, we formally describe TorchGeo, propose and test methods for sampling from large geospatial datasets, and test the effect of ImageNet pretraining versus random weight initialization on several benchmark datasets. We achieve close to state-of-the-art results on all experimental datasets, despite focusing only on creating simple and reproducible results to serve as baselines for future work to build on. We further find that ImageNet pre-training significantly improves spatial generalization performance in a land cover mapping task. We believe that these results are interesting in their own right, and that they highlight the importance of TorchGeo to the larger machine learning community. TorchGeo is open source and available on GitHub: \url{https://github.com/microsoft/torchgeo}.

\section{Design} \label{sec:design}

Remotely sensed imagery datasets are usually formatted as \textit{scenes}, i.e., tensors $X \in \mathbb{R}^{H \times W \times C}$, where $H$ is height, $W$ is width, and $C$ is the number of spectral channels, with corresponding spatial and temporal metadata. This metadata includes a \textit{coordinate reference system} (CRS) that maps pixel coordinates to the surface of the Earth, a \textit{spatial resolution} (the size of each pixel when mapped onto the surface of the Earth), \textit{spatial bounds} (a bounding box representing the area on Earth that the data covers), and a \textit{timestamp} or time range to indicate when the data was collected. We say that two datasets, $X^1$ and $X^2$, are \textit{pixel-aligned} if $X^1_{i,j}$ and $X^2_{i,j}$ represent data from the same positions on Earth for all $i,j$. Most pairs of datasets are not aligned by default. For example, $X^1$ and $X^2$ can be captured by two satellites in different orbits and will only have partially overlapping spatial bounds, or $X^1$ will be satellite imagery while $X^2$ will be from a dataset of labels with a different CRS (see Figure \ref{fig:warping-example}). However, deep learning model training requires pixel-aligned patches of imagery---i.e., smaller crops from large scenes. Most models are trained with mini-batch gradient descent and require input tensors in the format $B \times H \times W \times C$ where $B$ is the number of samples in a mini-batch and $W$ and $H$ are constant over all samples in the batch. At a higher level, training semantic segmentation models requires pairs of pixel-aligned imagery and masks.

Aligning two datasets requires \textit{reprojecting} the data from one file in order to match the CRS of the other file, \textit{cropping} the data to the same spatial bounds, and \textit{resampling} the data to correct for differences in resolution or to establish the same underlying pixel grid. Typically, these are performed as pre-processing steps using GIS software such as QGIS, ArcGIS, or tools provided by GDAL---see Listing~\ref{subsec:gdal_example} for an example of a GDAL command to align two data layers. This requires some level of domain knowledge to perform correctly and does not scale to large datasets as it requires creating duplicates of the layer to be aligned. Further, this approach still requires an implementation of a dataset or data loader that can sample patches from the pre-processed imagery.

In TorchGeo, we facilitate this process by performing the alignment logic on the fly to create pixel-aligned patches of data sampled from larger scenes. Specifically, we implement the alignment logic in custom PyTorch Dataset classes that are indexed in terms of spatiotemporal coordinates. Given a query in spatiotemporal coordinates, a desired destination CRS, and a desired spatial resolution, the custom dataset is responsible for returning the corresponding reprojected and resampled data for the query. We further implement geospatial data \textit{samplers} that generate queries according to different criteria (e.g., randomly or in a regular grid pattern). See Section~\ref{sec:implementation} for a discussion on the implementation of these.

As our datasets are indexed by spatiotemporal coordinates, we can easily compose datasets that represent different data layers by specifying a valid area to sample from. For example, if we have two datasets, $D^1$ and $D^2$, we may want to sample data from the union of the layers, $D^1 \cup D^2$, if both layers cover disparate geospatial locations, or from the intersection of the layers, $D^1 \cap D^2$, if one layer is imagery and the other layer is labels. The latter is particularly powerful for use in applications like multimodal learning~\citep{8269806} and data fusion~\citep{zhang2010multi}. As we describe in the following section, we implement generic dataset classes for a variety of common remotely-sensed datasets (e.g., Landsat imagery) that can be composed in this way. This allows users of the library to create their own multimodal datasets without having to write custom code.

Most importantly, these abstractions that TorchGeo creates---geospatial datasets and samplers---can be combined with a standard PyTorch data loader class to produce fixed size batches of data to be transferred to the GPU and used in training or inference. See Listing~\ref{listing:example} for a working example of TorchGeo code for setting up a functional data loader that uses Landsat and Cropland Data Layer (CDL) dataset implementations. Our approach trades off required storage space for data loading time compared to pre-processing all data layers, but, crucially, does not require knowledge of GIS tooling. We benchmark our implementations in Section~\ref{subsec:dataset_benchmarks}.

\section{Implementation} \label{sec:implementation}

The implementation of TorchGeo follows the design of other PyTorch domain libraries to reduce the amount of new concepts that a user must learn to integrate it with their existing workflow. We split TorchGeo up into the following submodules:
\begin{description}
    \item[Datasets] Our dataset implementations consist of both \textit{benchmark datasets} that allow users to interface with common datasets used in RS literature and \textit{generic datasets} that allow users to interface with common geospatial data layers such as Landsat or Sentinel-2 imagery. Either of these types of datasets can also be \textit{geospatial datasets}, i.e., datasets that contain geospatial metadata and can be sampled as such. These are part of the core contribution of TorchGeo and we describe them further in the following section.
    \item[Samplers] We implement samplers for indexing into any of our \textit{geospatial datasets}. Our geospatial datasets are indexed by bounding boxes in spatiotemporal coordinates (as opposed to standard fixed-length datasets of images which are usually indexed by an integer). The samplers generate bounding boxes according to specific patterns: randomly across all scenes in a dataset, random batches from single scenes at a time, or in grid patterns over scenes. Different sampling patterns can be useful for different model training strategies, or for running model inference over datasets. 
    \item[Models] Most existing model implementations (e.g., in torchvision) are fixed to accept 3 channel inputs which are not compatible with multispectral imagery. We provide implementations (or wrappers around well-established implementations) of common deep learning model architectures with variable-sized inputs and pre-trained weights, e.g., models that use all of the Sentinel-2 multispectral bands as inputs. We also implement architectures from recent geospatial ML work such as the Fully Convolutional Siamese Network~\citep{daudt2018fully}.
    \item[Transforms] Similar to existing model implementations, some existing deep learning packages do not support data augmentation methods for multi-spectral imagery. We provide wrappers for augmentations in the Kornia~\citep{eriba2019kornia} library (which does support augmentations over arbitrary channels) and implement transforms specific to geospatial data.
    \item[Trainers] Finally, we implement model training recipes using the PyTorch Lightning library~\citep{William_PyTorch_Lightning_2019}. These include both dataset-specific training code, and general training routines, e.g., an implementation of the BYOL self-supervision method~\citep{grill2020bootstrap}.
\end{description}

\subsection{Datasets}

\begin{table*}[t]
\centering
\resizebox{1.0\textwidth}{!}{%
\begin{tabular}{lccccccc}
\toprule
\textbf{Dataset} & \textbf{Task} & \textbf{Source} & \textbf{\# Samples} & \textbf{\# Categories} & \textbf{Size (px)} & \textbf{Resolution (m)} & \textbf{Bands} \\
\midrule \addlinespace
ADVANCE~\citep{10.1007/978-3-030-58586-0_5} & C & Google Earth, Freesound & 5,075 & 13 & $512\times512$ & 0.5 & RGB \\ \addlinespace
BigEarthNet~\citep{8900532} & C & Sentinel-1/2 & 590,326 & 19--43 & $120\times120$ & 10 & SAR, MSI \\ \addlinespace
EuroSAT~\citep{helber2019eurosat} & C & Sentinel-2 & 27,000 & 10 & $64\times64$ & 10 & MSI \\ \addlinespace
PatternNet~\citep{zhou2018patternnet} & C & Google Earth & 30,400 & 38 & $256\times256$ & 0.06--5 & RGB \\ \addlinespace
RESISC45~\citep{7891544} & C & Google Earth & 31,500 & 45 & $256\times256$ & 0.2--30 & RGB \\ \addlinespace
So2Sat~\citep{zhu2019so2sat} & C & Sentinel-1/2 & 400,673 & 17 & $32\times32$ & 10 & SAR, MSI \\ \addlinespace
UC Merced~\citep{yang2010bag} & C & USGS National Map & 21,000 & 21 & $256\times256$ & 0.3 & RGB \\ \addlinespace
COWC~\citep{mundhenk2016a} & C, R & UAV & 388,435 & 2 & $256\times256$ & 0.15 & RGB \\ \addlinespace
Tropical Cyclone~\citep{9149719} & R & GOES 8--16 & 108,110 & - & $256\times256$ & 4K--8K & MSI \\ \addlinespace
USAVars~\citep{rolf2021generalizable} & R & NAIP & 100K & - & - & 4 & MSI \\ \addlinespace
Benin Cashews~\citep{jin2021smallholder} & S & Airbus Pléiades & 70 & 6 & 1,$186\times1$,122 & 0.5 & MSI \\ \addlinespace
ETCI 2021 Floods~\citep{etci2021} & S & Sentinel-1 & 66,810 & 2 & $256\times256$ & 5--20 & SAR \\ \addlinespace
GID-15~\citep{tong2020land} & S & Gaofen-2 & 150 & 15 & 6,$800\times7$,200 & 3 & RGB \\ \addlinespace
Kenya Crop Type~\citep{radiant2020cv4a} & S & Sentinel-2 & 4,688 & 7 & 3,$035\times2$,016 & 10 & MSI \\ \addlinespace
LandCover.ai~\citep{boguszewski2021landcover} & S & Aerial & 10,674 & 5 & $512\times512$ & 0.25--0.5 & RGB \\ \addlinespace
Potsdam~\citep{rottensteiner2012isprs} & S & Aerial & 38 & 6 & 6,$000\times6$,000 & 0.05 & MSI \\ \addlinespace
SEN12MS~\citep{schmitt2019sen12ms} & S & Sentinel-1/2, MODIS & 180,662 & 33 & $256\times256$ & 10 & SAR, MSI \\ \addlinespace
Vaihingen~\citep{rottensteiner2012isprs} & S & Aerial & 33 & 6 & 1,281--3,816 & 0.09 & RGB \\ \addlinespace
FAIR1M~\citep{sun2021fair1m} & O & Gaofen, Google Earth & 15K & 37 & 1,$024\times1$,024 & 0.3--0.8 & RGB \\ \addlinespace
IDTReeS~\citep{graves2021data} & O, C & Aerial & 591 & 33 & $200\times200$ & 0.1--1 & RGB \\ \addlinespace
NWPU VHR-10~\citep{CHENG2014119} & O, I & Google Earth, Vaihingen & 800 & 10 & 358--1,728 & 0.08--2 & RGB \\ \addlinespace
SpaceNet~\citep{van2018spacenet} & I & WorldView, Planet Dove & 1,889--28,728 & 2 & 102--900 & 0.5--4 & MSI \\ \addlinespace
ZueriCrop~\citep{turkoglu2021crop} & I, T & Sentinel-2 & 116K & 48 & $24\times24$ & 10 & MSI \\ \addlinespace
Seasonal Contrast~\citep{manas2021seasonal} & T & Sentinel-2 & 100K--1M & - & $264\times264$ & 10 & MSI \\ \addlinespace
LEVIR-CD+~\citep{shen2021s2looking} & D & Google Earth & 985 & 2 & 1,$024\times1$,024 & 0.5 & RGB \\ \addlinespace
OSCD~\citep{daudt2018urban} & D & Sentinel-2 & 24 & 2 & 40--1,180 & 60 & MSI \\ \addlinespace
xView2~\citep{gupta2019xbd} & D & Maxar & 3,732 & 4 & 1,$024\times1$,024 & 0.8 & RGB \\ \addlinespace
\bottomrule \\[-0.15cm]
\multicolumn{8}{l}{C = classification,  R = regression, S = semantic segmentation, O = object detection, I = instance segmentation, T = time series, D = change detection}
\end{tabular}%
}
\caption{Benchmark datasets implemented in TorchGeo.}
\label{tab:benchmark-datasets}
\end{table*}

We organize datasets based on whether they are \textit{generic datasets} or \textit{benchmark datasets} and based on whether or not they contain geospatial metadata---i.e., are a \textit{geospatial dataset}.

Benchmark datasets are datasets released by the community that consist of both inputs and target labels for a specific type of task (scene classification, semantic segmentation, instance segmentation, etc.). These may or may not also contain geospatial metadata that allows them to be joined with other sources of data. In our opinion, one of the strongest components of existing deep learning domain libraries is the way that they make the use of existing datasets easy. We aim to replicate this and, for example, include options that let users automatically download the data for a corresponding dataset. Table~\ref{tab:benchmark-datasets} lists the set of benchmark datasets that TorchGeo currently supports.

\begin{table}[t]
\begin{tabular}{cc}
\hline
\textbf{Type}                          & \textbf{Dataset}                                                                                     \\ \toprule
\multirow{5}{*}{\textbf{Imagery}} & Landsat~\citep{roy2014landsat}                                                              \\
                              & Sentinel~\citep{drusch2012sentinel}                                                         \\
                              & NAIP~\citep{naip} \\
                              & ASTER Global DEM~\citep{abrams2020aster}                                                                            \\
                              & European DEM                                                                                \\ \midrule
\multirow{12}{*}{\textbf{Labels}} & Aboveground Woody Biomass~\citep{watch2002global}            \\
                              & Canadian Buildings Footprints~\citep{cbf}                                                   \\
                              & Chesapeake Land Cover~\citep{robinson2019large}                                             \\
                              & Global Mangrove Distribution~\citep{cms-global-mangrove}                \\
                              & Cropland Data Layer~\citep{boryan2011monitoring}                                            \\
                              & EDDMapS~\citep{bargeron2007eddmaps}   \\
                              & EnviroAtlas~\citep{enviroatlas}                                                                                 \\
                              & Esri 2020 Land Cover  ~\citep{esri2020}                                                                      \\
                              & GBIF~\citep{gbif}          \\
                              & GlobBiomass ~\citep{santoro2018ggdo}                                                                                \\
                              & iNaturalist~\citep{gbif}                                                                                 \\
                              & Open Buildings~\citep{sirko2021continental}                                                                              \\ \bottomrule
\end{tabular}
\caption{Generic datasets implemented in TorchGeo.}
\label{tab:generic-datasets}
\end{table}

Generic datasets are not created with a specific task in mind, but instead represent layers of geospatial data that can be used for any purpose. For example, we implement datasets for representing collections of scenes of Landsat imagery that let users index into the imagery and use it in arbitrary PyTorch-based pipelines. These are not limited to imagery; for example, we also implement a dataset representing the Cropland Data Layer labels, an annual raster layer that gives the estimated crop type or land cover at a 30~m/px resolution for the contiguous United States (see Table~\ref{tab:generic-datasets}).

\subsection{Samplers}

As our \textit{geospatial datasets} are indexed with bounding boxes using spatiotemporal coordinates and do not have a concept of a dataset ``length'', they cannot be sampled from by choosing a random integer. We provide three types of samplers for different situations: a RandomGeoSampler that returns a fixed-sized bounding box from the valid spatial extent of a dataset uniformly at random, a RandomBatchGeoSampler that returns a set of randomly positioned fixed-sized bounding boxes from a random scene within a dataset, and a GridGeoSampler that returns bounding boxes in a grid pattern over subsequent scenes within a dataset. These samplers are benchmarked in Section~\ref{subsec:dataloader_benchmarks}. This abstraction also allows for methods that rely on specific data sampling patterns. For example, Tile2Vec~\citep{jean2019tile2vec} relies on sampling triplets of imagery where two of the images are close to each other in space while the third is distant. This logic can be implemented in several lines of code as a custom sampler class, that would then operate over any of the generic imagery datasets. Finally, all TorchGeo samplers are compatible with PyTorch data loader objects, allowing them to be fit into any PyTorch-based pipeline.

\section{Experiments and Results}

\subsection{Datasets} \label{subsec:datasets}

We use the following datasets in our experiments:

\begin{description}
\item[Landsat and CDL] A collection of multispectral imagery from 114 Landsat~8~\citep{roy2014landsat} scenes and the Cropland Data Layer (CDL)~\citep{boryan2011monitoring} dataset. This data is 151~GB on disk and is stored in cloud optimized GeoTIFF (COG) format. We use this dataset to benchmark our GeoDataset and sampler implementations.
\item[So2Sat] A \textit{classification} dataset consisting of 400,673 image patches classified with one of 42 local climate zone labels~\citep{zhu2019so2sat}. The patches are $30 \times 30$ pixels in size with 18 channels consisting of Sentinel-1 and Sentinel-2 bands. They are sampled from different urban areas around the globe. We use the second version of the dataset as described on the project's GitHub page\footnote{\url{https://github.com/zhu-xlab/So2Sat-LCZ42}} in which the training split consists of data from 42 cities around the world, the validation split consists of the western half of 10 other cities, and the testing split covers the eastern half of the 10 remaining cities.
\item[LandCover.ai] A \textit{semantic segmentation} dataset consisting of high-resolution (0.5~m/px and 0.25~m/px) RGB aerial imagery from 41 tiles over Poland where each pixel has been classified as one of five land cover classes~\citep{boguszewski2021landcover}. The scenes are divided into 10,674 $512 \times 512$ pixel patches and split according to the script on the dataset webpage\footnote{\url{https://landcover.ai}}.
\item[Chesapeake Land Cover] A \textit{semantic segmentation} dataset consiting of high-resolution (1~m/px) imagery from the USDA's National Agriculture Imagery Program (NAIP) and high-resolution (1~m/px) 6-class land cover labels from the Chesapeake Conservancy~\citep{robinson2019large}. The dataset contains imagery and land cover masks for parts of six states in the Northeastern US. The data for each state is split into $\sim7\times6$ km tiles, then divided into pre-defined train, validation, and test splits\footnote{\url{https://lila.science/datasets/chesapeakelandcover}}.
\item[RESISC45] A \textit{classification} dataset consisting of 31,500 $256 \times 256$ pixel RGB image patches of varying spatial resolutions where each patch is classified into one of 45 classes~\citep{zhang2019remote}. As the dataset does not have official splits, we use the train/val/test splits defined in~\citet{neumann2019domain}.
\item[ETCI 2021] A \textit{semantic segmentation} dataset used in a flood detection competition~\citep{etci2021}. It consists of 66,810 $256\times256$ pixel Sentinel-1 SAR images. We use the official train/test splits, and we further randomly subdivide the train split 80/20 into train/val splits.
\item[EuroSAT] A \textit{classification} dataset consisting of 27,000 $64\times64$ pixel Sentinel-2 images and 10 target classes~\citep{helber2019eurosat}. We use the train/val/test splits defined in~\citet{neumann2019domain}.
\item[UC Merced] A \textit{classification} dataset consisting of 21,000 $256\times256$ pixel RGB images from the USGS National Map and 21 target classes~\citep{yang2010bag}. We use the train/val/test splits defined in~\citet{neumann2019domain}.
\item[COWC Counting] A \textit{regression} dataset consisting of 317,230 training and 81,161 testing images~\citep{mundhenk2016a}. Each $256\times256$ pixel image is labeled with the number of cars in the image. We reserved 81,161 images from the training set for validation.
\end{description}

\subsection{Data loader benchmarks} \label{subsec:dataloader_benchmarks}

\begin{figure*}[t]
    \centering
    \begin{subfigure}[b]{0.45\textwidth}
        \caption{Sampler performance vs. batch size}
        \includegraphics[width=\textwidth]{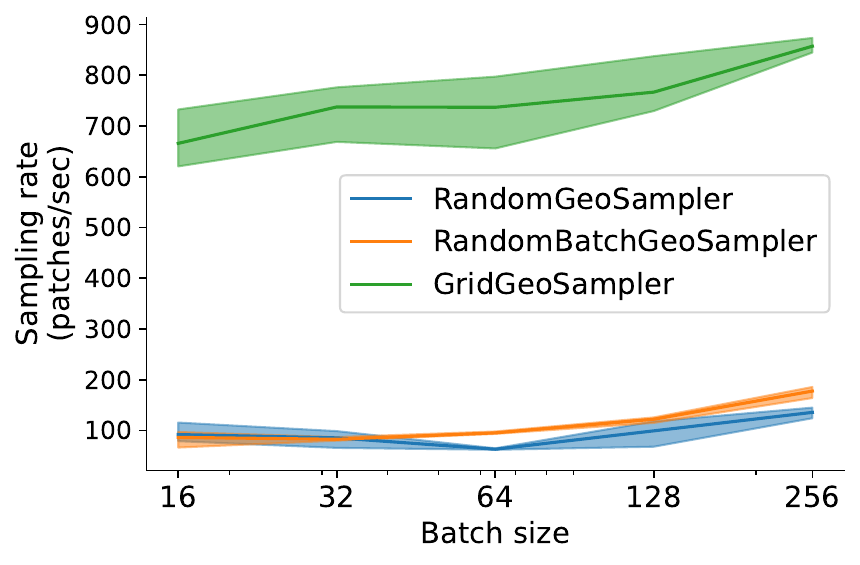}
        \label{fig:load}
    \end{subfigure}
    ~
    \begin{subfigure}[b]{0.45\textwidth}
        \caption{Effect of preprocessing and caching}
        \includegraphics[width=\textwidth]{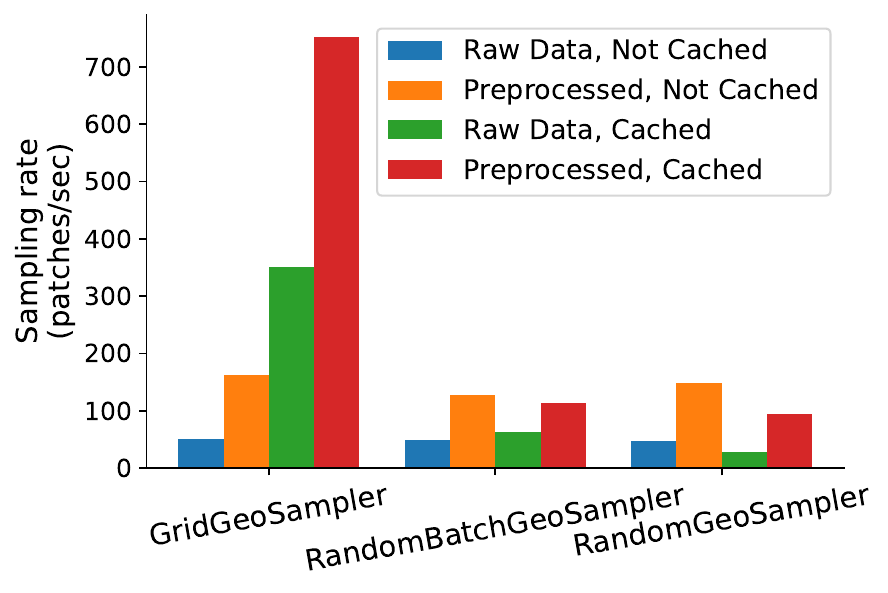}
        \label{fig:warp}
    \end{subfigure}
    \caption{Sampling performance of various GeoSampler implementations. (a) Solid lines represent average sampling rate, while shaded region represents minimum and maximum performance across random seeds. (b) Average sampling rate under different data loading conditions.}\label{fig:dataloader}
\end{figure*}

We first benchmark the speed at which TorchGeo can sample patches of imagery and masks from the Landsat and CDL dataset. We believe this dataset is typical of a large class of geospatial machine learning problems---where users have access to a large amount of satellite imagery scenes covering a broad spatial extent and, separately, per-pixel label masks where each scene is not necessarily projected in the same coordinate reference system. The end goal of such a problem is to train a model with pixel-aligned patches of imagery and label masks as described in Section~\ref{sec:design}. As such, we measure the rate at which our dataset and sampler implementations can provide patches to a GPU for training and inference.

In Figure~\ref{fig:load}, we calculate the rate at which samples of varying batch size can be drawn from a GeoDataset using various GeoSampler implementations. Compared to the other samplers, GridGeoSampler is significantly faster due to the repeated access of samples that are already in GDAL's least recently used (LRU) cache. For small batch sizes, RandomGeoSampler and RandomBatchGeoSampler are almost identical, since overlap between patches is uncommon. However, for larger batch sizes, RandomBatchGeoSampler starts to outperform RandomGeoSampler as the cache is used more effectively. 

In Figure~\ref{fig:warp}, we demonstrate the difference that preprocessing and caching data makes. This is most easily demonstrated by GridGeoSampler and to a lesser extent the other samplers. GDAL's LRU cache only saves raw data loading times, so warping must always be done on the fly if the dataset CRSs or resolutions do not match. When the necessary storage is available, preprocessing the data ahead of time can lead to significantly faster sampling rates. Although RandomGeoSampler and RandomBatchGeoSampler are much slower than GridGeoSampler, most users will only need to use GridGeoSampler for inference due to our pre-trained model weights.

\subsection{Dataset benchmarks} \label{subsec:dataset_benchmarks}

\begin{table*}[t]
\centering
\begin{tabular}{@{}ccccc@{}}
\toprule
\textbf{Dataset} & \textbf{Method} & \textbf{Weight Initialization} & \textbf{Bands} & \textbf{Performance} \\ \midrule
\multirow{5}{*}{RESISC45~\citep{7891544}} & ResNet50 & ImageNet & RGB & 95.42 $\pm$ 0.23\% \\
 & ResNet18 & random & RGB & 79.90 $\pm$ 0.25\% \\
 & ResNet50 v2~\citep{neumann2019domain} & In domain & RGB & \textbf{96.86\%} \\
 & ViT B/16~\citep{steiner2021train} & ImageNet-21k & RGB & 96.80\% \\
 & ResNet50~\citep{zhai2019large} & Sup-Rotation-100\% & RGB & 96.30\% \\ \midrule
\multirow{6}{*}{So2Sat~\citep{zhu2019so2sat}} & ResNet50 & ImageNet (+ random) & MSI & \textbf{63.99 $\pm$ 1.38\%} \\
 & ResNet50 & random & MSI & 56.82 $\pm$ 4.32\% \\
 & ResNet50 & ImageNet & RGB & 59.82 $\pm$ 0.94\% \\
 & ResNet50 & random & RGB & 49.46 $\pm$ 2.67\% \\
 & ResNeXt + CBAM~\citep{zhu2019so2sat} & random & MSI & 61\% \\
 & ResNet50 v2~\citep{neumann2019domain} & In domain & RGB & \textbf{63.25\%} \\ \midrule
\multirow{3}{*}{LandCover.ai~\citep{boguszewski2021landcover}} & U-Net, ResNet50 encoder & ImageNet & RGB & 84.81 $\pm$ 0.21\% \\
 & U-Net, ResNet50 encoder & random & RGB & 79.73 $\pm$ 0.67\% \\
 & \begin{tabular}[c]{@{}c@{}}DeepLabv3+, Xception71 with\\ DPC encoder~\citep{boguszewski2021landcover}\end{tabular} & Cityscapes & RGB & \textbf{85.56\%} \\ \midrule
\multirow{2}{*}{\begin{tabular}[c]{@{}c@{}}Chesapeake Land Cover~\citep{robinson2019large}\\ Delaware split\end{tabular}} & U-Net, ResNet50 encoder & ImageNet (+ random) & MSI & \textbf{69.40 $\pm$ 1.39\%} \\
 & U-Net, ResNet18 encoder & random & MSI & \textbf{68.99 $\pm$ 0.84\%} \\
\midrule
ETCI 2021~\citep{etci2021} & U-Net, ResNet50 encoder & random & SAR & \textbf{45.77 $\pm$ 3.19\%} \\
\midrule
\multirow{5}{*}{EuroSAT~\citep{helber2019eurosat}} & ResNet50 & ImageNet (+ random) & MSI & 97.86 $\pm$ 0.23\% \\
 & ResNet50 & random & MSI & 96.07 $\pm$ 0.28\% \\
 & ResNet50 & ImageNet & RGB & 98.11 $\pm$ 0.31\% \\
 & ResNet50 & random & RGB & 87.33 $\pm$ 0.76\% \\
 & ResNet50 v2~\citep{neumann2019domain} & In domain & RGB & \textbf{99.20\%} \\
\midrule
\multirow{2}{*}{UC Merced~\citep{yang2010bag}} & ResNet50 & ImageNet & RGB & 98.15\% $\pm$ 0.46\% \\
 & ResNet50 v2~\citep{neumann2019domain} & In domain & RGB & \textbf{99.61\%} \\
\midrule
\multirow{3}{*}{COWC Counting~\citep{mundhenk2016a}} & ResNet50 & ImageNet & RGB & \textbf{0.573 $\pm$ 0.005} \\
 & ResNet18 & ImageNet & RGB & 0.667 $\pm$ 0.007 \\
 & ResCeption~\citep{mundhenk2016a} & random & RGB & 0.657 \\
\bottomrule
\end{tabular}
\caption{Benchmark results comparing TorchGeo trained models to previously reported results over 8 datasets. Classification dataset results are reported as overall top-1 accuracy, semantic segmentation dataset results are reported as mean class IoU, and regression dataset results are reported as RMSE. Results from TorchGeo models are reported as the mean with one standard deviation over 10 \textit{training} runs from different random seeds. Results from related work are reported as is.}
\label{tab:main-results}
\end{table*}

We also use TorchGeo to create simple, reproducible benchmark results on 8 of the datasets described in Section \ref{subsec:datasets}. We report the uncertainty in the results such that future work can evaluate whether a proposed improvement is due to methodological innovation or variance in results due to the stochastic nature of training deep learning models. To ensure reproducibility, we include a model training and evaluation framework in TorchGeo based around the PyTorch Lightning library and release all of our results. To quantify uncertainty, we report the mean and standard deviation metrics calculated over 10 training runs with different random seeds. The torchmetrics library~\citep{torchmetrics} was used to compute all performance values (overall top-1 accuracy for the classification datasets, mean intersection over union (mIoU) for the semantic segmentation datasets, and root mean square error (RMSE) for the regression dataset).

Our main results are shown in Table \ref{tab:main-results}. We find that our simple training setup achieves competitive results on several datasets. The in-domain pre-training method in~\citet{neumann2019domain} trains models starting from ImageNet weights, then further trains on remote sensing (in domain) datasets, before fine-tuning on the actual target dataset, which performs better than simply starting from ImageNet weights. In contrast, our best result across the RESISC45, EuroSAT, and UC Merced datasets come from simply using ImageNet pre-trained models and training on the target dataset with a low learning rate. The difference between these approaches is likely entirely due to different learning rate selection in the hyperparameter search, and data augmentation in the in-domain pre-training setup. On the So2Sat dataset, we find that the ImageNet pre-trained models are able to achieve similar results as in-domain pretraining, but only when using all Sentinel-2 bands. The previously reported baseline methods on the LandCover.ai dataset all use a DeepLabV3+ segmentation model with a Xception71 + Dense Prediction Cell (DPC) encoder that has been pre-trained on Cityscapes. We are able to achieve a result within 0.75\%~mIoU of this setup using a simple U-Net and ResNet50 encoder pre-trained on ImageNet. Finally, we report the first set of basic benchmark results on the Chesapeake Land Cover dataset.

\subsection{Effect of ImageNet pre-training on generalization performance}

Two of the datasets we test with, So2Sat and Chesapeake Land Cover, contain splits that are designed to measure the generalization performance of a model. The validation and test splits from So2Sat include data from urban areas that are not included in the training split while the Chesapeake Land Cover dataset contains separate splits for six different states. In these settings, we observe a large performance boost when training models from ImageNet weights versus from a random initialization, however we do not observe the boost on in-domain data. Table~\ref{tab:generalization} shows the performance of models that are trained on the Delaware split and evaluated on the test splits from every state. The in-domain performance (i.e., in Delaware) of the ImageNet pre-trained model and a model trained from scratch are the same, however, in every other setting the ImageNet pre-trained model performs better. In all cases but Maryland, this difference is greater than 6 points of mIoU with the most extreme difference being 12 points in Virginia. In the So2Sat case, we find that most models are not able to significantly reduce validation loss (where in this case validation data is out-of-domain), while, unsurprisingly, achieving near perfect results on the training data. Despite this overfitting, there remains a large gap between the best and worst models, with ImageNet pre-trained models achieving +7\% and +10\% accuracy over randomly initialized models. These results extend existing lines of research in computer vision that show how pre-training can improve out-of-domain performance~\citep{hendrycks2019using}, and how, specifically, ImageNet pretraining is useful for transfer learning tasks~\citep{huh2016makes}. 

\begin{table*}[t]
\centering
\resizebox{\textwidth}{!}{%
\begin{tabular}{@{}ccccccc@{}}
\toprule
\textbf{Weight init} & \textbf{Delaware} & \textbf{Maryland} & \textbf{New York} & \textbf{Pennsylvania} & \textbf{Virginia} & \textbf{West Virginia} \\ \midrule
ImageNet (+ random)  & \textbf{69.40 $\pm$ 1.39\%}  & \textbf{59.57 $\pm$ 0.70\%}  & \textbf{57.95 $\pm$ 1.10\%}  & \textbf{55.13 $\pm$ 1.25\%}      & \textbf{45.56 $\pm$ 1.54\%}  & \textbf{20.76 $\pm$ 1.95\%}       \\
random                 & 68.99 $\pm$ 0.84\%  & 57.30 $\pm$ 0.78\%  & 49.26 $\pm$ 2.40\%  & 47.67 $\pm$ 2.40\%      & 33.14 $\pm$ 3.73\%  & 14.95 $\pm$ 2.72\%       \\ \bottomrule
\end{tabular}%
}
\caption{Mean IoU performance of models trained in Delaware, with and without ImageNet weight initialization, on the test splits from Chesapeake Land Cover dataset.}
\label{tab:generalization}
\end{table*}

\section{Discussion}

We introduce TorchGeo, a Python package for enabling deep learning with geospatial data. TorchGeo provides data loaders for common geospatial datasets, composable data loaders for uncurated geospatial raster and vector data, samplers appropriate for geospatial data, models pre-trained on satellite imagery, multispectral transforms, and model trainers. Importantly, TorchGeo allows users to bypass the common pre-processing steps necessary to align geospatial imagery with labels and performs this pre-processing on the fly. We benchmark TorchGeo data loader speed and demonstrate how TorchGeo can be used to create reproducible benchmark results in several geospatial datasets.

Finally, TorchGeo serves as a platform for performing geospatial machine learning research. Existing works in self-supervised learning with geospatial data rely on spatiotemporal metadata and can be naturally implemented in TorchGeo and scaled over large amounts of geospatial imagery without the need for pre-processing steps. Similarly, augmentation methods appropriate for training with geospatial imagery are under-explored, however, can be easily integrated with TorchGeo. Other interesting future research directions include building inductive biases appropriate for geospatial imagery into deep learning models (similar to the work done on rotation equivariant networks~\citep{marcos2018land}), data fusion techniques (e.g., how to incorporate spatial information into models, or appropriately use multi-modal layers), and learning shape-based models. Finally, TorchGeo exposes a catalog of benchmark geospatial datasets (Table \ref{tab:benchmark-datasets}) through a common interface, and, with the results in this paper, has begun to include corresponding benchmark results. This makes it easy for researchers to compare new ideas to existing work without having to repeat expensive computations. We hope TorchGeo can help drive advances at the intersection of machine learning and remote sensing.

\begin{acks}
This research is part of the Blue Waters sustained-petascale computing project, which is supported by the National Science Foundation (awards OCI-0725070 and ACI-1238993), the State of Illinois, and as of December, 2019, the National Geospatial-Intelligence Agency. Blue Waters is a joint effort of the University of Illinois at Urbana-Champaign and its National Center for Supercomputing Applications. The research was supported by NSF grants IIS 21-31335, OAC 21-30835, DBI 20-21898, and a C3.ai research award. We'd like to thank TorchGeo contributors for their efforts in creating the library. We'd also like to thank Siyu Yang, Md Nasir, Shahrzad Gholami, and Thomas Roca for their feedback and support.
\end{acks}

\bibliographystyle{ACM-Reference-Format}
\bibliography{citations}


\appendix

\section{Data loader benchmarking}

In order to benchmark the performance of our data loaders, we download 114 Landsat 8 collection 2 level-2 scenes and 1 Cropland Data Layer (CDL) file for the year of 2019. All files are stored as Cloud Optimized GeoTIFFs (COGs) with a block size of 512, and take up 151~GB of space on disk, uncompressed. All files are kept in their original CRS (Albers Equal Area for CDL and UTM for Landsat). Experiments are run on Microsoft Azure with a 6-core Intel Xeon E5-2690 CPU. All data is stored on a local SSD attached to the compute node. Batch size and random seed are varied while the remaining hyperparameters are kept fixed. Total epoch size is 4096, patch size is 224, stride is 112, and the number of workers for parallel data loading is set to 6.

\section{Pre-processing alignment with GDAL} 

As an example of an alignment pre-processing workflow, we assume that we have a Landsat 8 scene and a Cropland Data Layer (CDL) raster (``cdl.tif'') which completely covers the extent of the Landsat scene. We would like to create a pixel-aligned version of these two layers. Given that the Landsat 8 scene has a CRS of ``EPSG:32619'', a height of 8011 pixels, a width of 7891 pixels, and spatial bounds of (186585, 4505085, 423315, 4745415), the corresponding GDAL command to create a cropped version of the CDL layer that is aligned to the Landsat layer would look like:

\definecolor{background}{RGB}{255, 255, 255}
\definecolor{string}{RGB}{0, 18, 166}
\definecolor{comment}{RGB}{117, 113, 94}
\definecolor{normal}{RGB}{0, 0, 0}
\definecolor{identifier}{RGB}{166, 226, 46}

\lstset{
  language=python,                		
  numbers=left,                   		
  stepnumber=1,                   		
  numbersep=7pt,                  		
  numberstyle=\tiny\color{black}\ttfamily,
  backgroundcolor=\color{background},  	
  showspaces=false,               		
  showstringspaces=false,         		
  showtabs=false,                 		
  tabsize=4,                      		
  captionpos=b,                   		
  breaklines=true,                		
  breakatwhitespace=true,         		
  title=\lstname,                 		
  basicstyle=\color{normal}\ttfamily,		
  keywordstyle=\color{magenta}\ttfamily,	
  stringstyle=\color{string}\ttfamily,	
  commentstyle=\color{comment}\ttfamily,	
  emph={format_string, eff_ana_bf, permute, eff_ana_btr},
  emphstyle=\color{identifier}\ttfamily
}

\begin{lstlisting}[language=sh, caption=Command-line example of manual reprojection of CDL using GDAL., label={subsec:gdal_example}]
$ gdalwarp \
    -t_srs EPSG:32619 \
    -of COG \ 
    -te 186585 4505085 423315 4745415 \ 
    -ts 7891 8011 \
    cdl.tif aligned_cdl.tif
\end{lstlisting}

The spatial metadata of the Landsat scene can be determined through other GDAL command-line tools (gdalinfo command), geospatial data packages such as the rasterio package in Python, or through GIS software such as QGIS or ArcGIS.

\section{TorchGeo code example}

\begin{lstlisting}[language=Python, caption=Example TorchGeo code for creating a joint Landsat 7/8~\citep{roy2014landsat} and Cropland Data Layer (CDL)~\citep{boryan2011monitoring} dataset and using such a dataset with a standard PyTorch DataLoader class., label={listing:example}, float]
from torch.utils.data import DataLoader
from torchgeo.datasets import (
    Landsat7, Landsat8, CDL, stack_samples
)
from torchgeo.samplers import (
    RandomGeoSampler
)

# Take the union of all Landsat imagery
landsat7 = Landsat7(root="...")
landsat8 = Landsat8(
    root="...", 
    bands=Landsat8.all_bands[1:-2]],
)
landsat = landsat7 | landsat8

# Take the intersection of Landsat and CDL
cdl = CDL(
    root="...", download=True, checksum=True
)
dataset = landsat & cdl

# Sample 10,000 256x256 image patches
# from the intersection of the datasets
sampler = RandomGeoSampler(
    dataset, size=256, length=10000
)

# Use the dataset and sampler as normal 
# in a PyTorch DataLoader
dataloader = DataLoader(
    dataset, 
    batch_size=128, 
    sampler=sampler, 
    collate_fn=stack_samples
)
for batch in dataloader:
    image = batch["image"]
    mask = batch["mask"]
    
    # Train a model, or make predictions 
    # using a pre-trained model
\end{lstlisting}

TorchGeo is designed to be simple and easy to use, and provides a familiar API for users who have experience using libraries like torchvision~\citep{torchvision}. In Listing~\ref{listing:example}, we provide an example code snippet showing how to use TorchGeo.

In this example, we show how easy it is to work with geospatial data and to sample small image patches from a combination of Landsat~\citep{roy2014landsat} and Cropland Data Layer (CDL)~\citep{boryan2011monitoring} data using TorchGeo. First, we assume that the user has Landsat 7 and 8 imagery downloaded. Since Landsat 8 has more spectral bands than Landsat 7, we only use the bands that both satellites have in common. We create a single dataset including all images from both Landsat 7 and 8 data by taking the union between these two datasets.

Next, we take the intersection between this dataset and the CDL dataset. We want to take the intersection instead of the union to ensure that we only sample from regions where we have both Landsat and CDL data. Note that we can automatically download and checksum CDL data. Also note that each of these datasets may contain files in different CRSs or resolutions, but TorchGeo automatically ensures that a matching CRS and resolution is used.

This dataset can now be used with a PyTorch data loader. Unlike benchmark datasets, geospatial datasets often include very large images. For example, the CDL dataset consists of a single image covering the entire contiguous United States. In order to sample from these datasets using geospatial coordinates, TorchGeo defines a number of samplers. In this example, we use a random sampler that returns $256 \times 256$ pixel images and 10,000 samples per epoch. We also use a custom collation function to combine each sample dictionary into a mini-batch of samples. This data loader can now be used in your normal training/evaluation pipeline.

Many applications involve intelligently composing datasets based on geospatial metadata like this. For example, users may want to:

\begin{itemize}
    \item Combine datasets for multiple image sources and treat them as equivalent (e.g., Landsat 7 and 8)
    \item Combine datasets for disparate geospatial locations (e.g., Chesapeake NY and PA)
\end{itemize}

These combinations require that all queries are present in at least one dataset, and can be created using a UnionDataset. Similarly, users may want to:

\begin{itemize}
    \item Combine image and target labels and sample from both simultaneously (e.g., Landsat and CDL)
    \item Combine datasets for multiple images sources for multimodal learning or data fusion (e.g., Landsat and Sentinel)
\end{itemize}

These combinations require that all queries are present in both datasets, and can be created using an IntersectionDataset. TorchGeo automatically composes these datasets for you when you use the intersection (\texttt{\&}) and union (\texttt{|}) operators.

\section{Benchmark dataset experiments}

For experiments, we use the pre-defined training, validation, and testing splits in all datasets. We perform a small hyperparameter search with a single fixed random seed on each dataset. Specifically, we perform a hyperparameter search for the best validation performance over the grid: model architecture $\in$ \{ResNet18, ResNet50\}, weight initialization $\in$ \{random, ImageNet\}, learning rate $\in$ \{0.01, 0.001, 0.0001\}, and loss function $\in$ \{cross entropy, jaccard\}. For the classification and regression datasets, we use the ResNets as is, and for the semantic segmentation datasets, we use the ResNets as the encoder model in a U-Net. With the So2Sat and EuroSAT datasets we also run experiments that use all the Sentinel-2 bands vs. only the RGB bands. In the cases where we use ImageNet weights with imagery that has more than RGB bands we randomly initialize the non-RGB kernels in the first convolutional layer of the network and denote this setting as ImageNet (+ random). In all cases, we use the AdamW optimizer~\citep{loshchilov2017decoupled}, reduce learning rate on validation loss plateaus, and early stop based on validation loss. We repeat the training process with 10 different seeds using the best performing hyperparameter configuration and report the test set performance over these models.

\end{document}